# Image Classification of Melanoma, Nevus and Seborrheic Keratosis by Deep Neural Network Ensemble

Kazuhisa Matsunaga[1]   Akira Hamada[1]   Akane Minagawa[2]   Hiroshi Koga[2]

**Abstract**

This short paper reports the method and the evaluation results of Casio and Shinshu University joint team for the ISBI Challenge 2017 – *Skin Lesion Analysis Towards Melanoma Detection* – Part 3: Lesion Classification hosted by ISIC. Our online validation score was 0.958 with melanoma classifier AUC 0.924 and seborrheic keratosis classifier AUC 0.993.

## 1. Introduction

The defined task of the Part 3 of the ISIC-ISBI 2017 Challenge *Skin Lesion Analysis Towards Melanoma Detection* [1] was to classify skin lesion images into three classes – melanoma (MM; malignant melanoma), nevus (NCN; nevocellular nevus) and seborrheic keratosis (SK) through two binary classifiers – MM vs. rest (MM classifier) and SK vs. rest (SK classifier). Participants were ranked by the mean value of the two classifier area under the ROC curves (AUCs). The regulation of this task allowed the use of external training data and the use of the age/sex information tagged with a number of the provided samples.

In this report, we show our proposed method – the classification system (section 2) and the machine learning (section 3) – and its evaluation results (section 4).

## 2. Proposed Classification System

Figure 1 shows our proposed classification system. First, luminance and color balance of input images are normalized exploiting color constancy [2]; the example results are shown in Figure 3. Normalized images are input to a base classifier trained for SK vs. rest as well as to a base classifier trained for MM vs. rest. Both base classifiers have identical composition as in Figure 2; geometrically transformed images (combinations of rotation, translation, scaling and flipping) are input in parallel to an ensemble of convolutional neural networks (CNNs) and a prediction value in [0.0,1.0] is output. We adopted 50-layer ResNet [3] implemented in Keras [4] with small modifications by ourselves.

Before the output of the prediction, age/sex information may be checked here. As seen in the training set, SK and MM are rare at young ages in general. We adopted a straightforward thresholding by age/sex information only for SK classification. For MM classification, we observed no significant increase by our cross-validation.

In addition, we noticed that our SK classifier were far more reliable than our MM classifier. The boundary of the SK class must be represented better with the SK classifier than with the MM classifier although they learned from the same sample set, so we integrated the information also into the MM classification. That is, if a sample is very likely to be SK, it probably won't be MM. We applied *ad hoc* linear approximation as follows:

$$F_{\text{MM}}(x) = \max\{0, \tilde{F}_{\text{MM}}(x) - \tilde{C}_{\text{MM}} - \alpha(F_{\text{SK}}(x) - C_{\text{SK}})\} \text{ (if } F_{\text{SK}}(x) > C_{\text{SK}}),$$

with $F(x)$: a classifier output for a sample $x$, $C$: a threshold at equal error rate (EER), and a tilde indicates a base classifier. We propose $\alpha = \frac{\tilde{C}_{\text{MM}}}{C_{\text{SK}}}$, for example.

---

[1] R&D Center, Casio Computer Co., Ltd., Japan, e-mail: `matsunagak` at `casio.co.jp`
[2] SHINSHU University, Japan

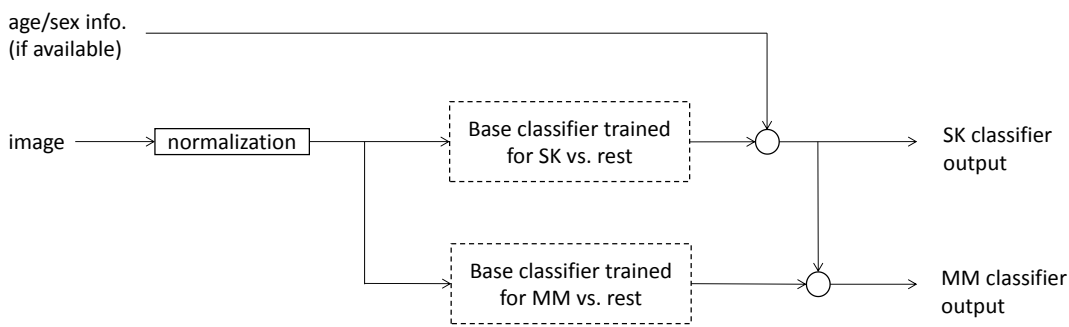

**Figure 1**

Proposed classification system

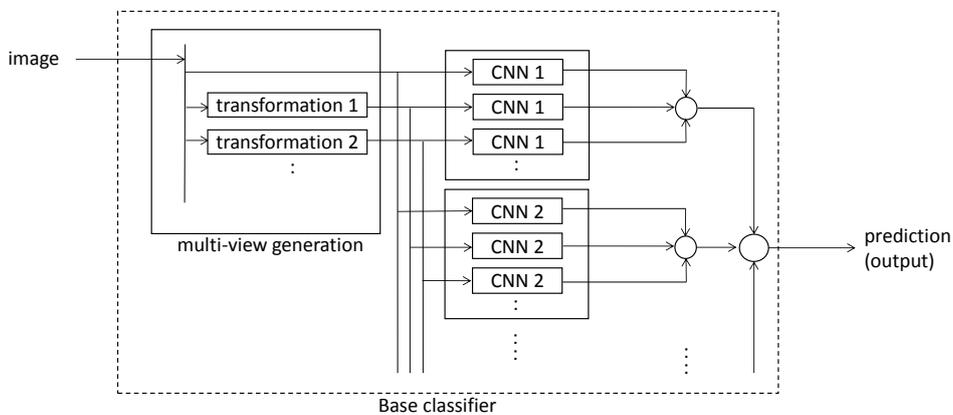

**Figure 2**

A base classifier composition

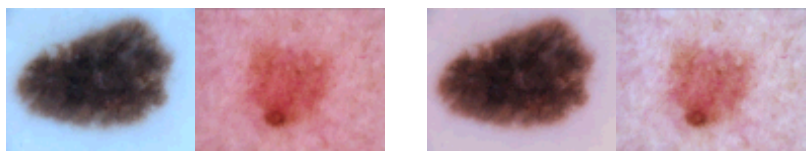

**Figure 3**

Examples of luminance and color normalization (left: original, right: normalized)

## 3. Machine Learning

In addition to the provided training data (374 MM, 254 SK, 1372 NCN samples), we used external training data (409 MM, 66 SK, 969 NCN samples) from the subset of the ISIC Archive [5].

Our CNNs were fine-tuned with the training samples from the initial pre-trained model for generic object recognition in Keras [4]. We applied different types of optimization and selected the best combination of fine-tuned CNNs through cross-validations. The optimization methods used were RMSProp [6] and AdaGrad [7].

## 4. Results

We show our evaluation results. Participants of the Challenge 2017 could check the online validation scores with the provided validation data (150 samples in total). Table 1 shows the final results of our proposed method.

Addition of external data improved SK AUC from 0.981 to 0.992 and MM AUC from 0.896 to 0.899 in cross-validation. As for age/sex information, we observed SK AUC improved from 0.957 to 0.960 in cross-validation and no improvement for MM. The complementary use of SK classifier for MM classification improved MM AUC from 0.917 to 0.924 for the validation set score.

Table 2 summarizes the results for the Challenge 2016 Part 3 which was malignant vs. benign binary classification. Ours was also trained using the 2016 training set with external samples from the ISIC archive (we carefully excluded the samples in the test set). The method published later in 2016 by Codella et al. [8] utilized a deep neural network-based automatic lesion segmentation process which ours does not have.

|  | Proposed |
|---|---|
| MM classifier AUC | 0.924 |
| SK classifier AUC | 0.993 |
| Mean | 0.958 |

**Table 1**

The ISIC-ISBI Challenge 2017 Part 3 validation set scores

|  | Proposed | Top of 2016 [9] | [8] without segmentation | Best of [8] with segmentation |
|---|---|---|---|---|
| AUC | 0.874 | 0.804 | 0.808 | 0.838 |
| Average precision | 0.744 | 0.637 | 0.596 | 0.645 |

**Table 2**

The ISIC-ISBI Challenge 2016 Part 3 test set scores

## 5. Conclusions and Future Work

Our proposed method significantly outperformed last year's state-of-the-art (Table 2), even though we could not utilize any lesion segmentation (or cropping) this time. As stated in [8], we expect that our results will improve to a certain extent if some reliable segmentation method is given.

On utilizing age/sex information for SK classification, we observed a slight effect. From a clinical point of view, one may need more careful implementation than thresholding in practical use, because SK at young ages are rare, but still possible [10].

The multi-class integration stated in the last of section 2 may require more mathematical sophistication and consideration of how to treat complex diseases.

## Acknowledgements

Authors would like to thank Prof. Masaru Tanaka (Tokyo Women's Medical University Medical Center East) and Dr. Toshitsugu Sato (Sato Dermatology Clinic, Tokyo) for great help and advice on our early prototype.